\documentclass[%
 reprint,
 superscriptaddress,
showpacs,
 amsmath,amssymb,
 aps,
 pre,
floatfix,
]{revtex4-1}

\usepackage{graphicx}
\usepackage{dcolumn}
\usepackage{bm}

\usepackage[english]{babel}
\usepackage{color}

\begin{document}

\preprint{APS/123-QED}

\title{The scarcity of crossing dependencies: \\ a direct outcome of a specific constraint?}

\author{Carlos G\'{o}mez-Rodr\'{i}guez}
 \affiliation{Universidade da Coru\~na\\
  FASTPARSE Lab, LyS Research Group\\
  Departamento de Computaci\'on \\
  Universidade da Coru\~{n}a \\
  Campus de Elvi\~{n}a, s/n \\
  15071 A Coru\~{n}a, Spain.}
\email{carlos.gomez@udc.es}
\homepage{http://www.grupolys.org/~cgomezr/}
\author{Ramon Ferrer-i-Cancho}
 \affiliation{Complexity and Quantitative Linguistics Lab \\
  LARCA Research Group \\
  Departament de Ci\`encies de la Computaci\'o \\
  Universitat Polit\`ecnica de Catalunya (UPC) \\
  Campus Nord, Edifici Omega\\
  08034 Barcelona, Catalonia, Spain.
 }
 \email{rferrericancho@cs.upc.edu}
 \homepage{http://www.cs.upc.edu/~rferrericancho/} 

\date{\today}

\begin{abstract}
The structure of a sentence can be represented as a network where vertices are words and edges indicate syntactic dependencies. Interestingly, crossing syntactic dependencies have been observed to be infrequent in human languages.
This leads to the question of whether the scarcity of crossings in languages arises from an independent and specific constraint on crossings.
We provide statistical evidence suggesting that this is not the case, as the proportion of dependency crossings of sentences from a wide range of languages can be accurately estimated by a simple predictor based on a null hypothesis on the local probability that two dependencies cross given their lengths. 
The relative error of this predictor never exceeds $5\%$ on average, whereas the error of a baseline predictor assuming a random ordering of the words of a sentence is at least 6 times greater.
Our results suggest that the low frequency of crossings in natural languages is neither originated by hidden knowledge of language nor by the undesirability of crossings {\em per se}, but as a mere side effect of the principle of dependency length minimization.
\end{abstract}

\maketitle



\section{Introduction}

\label{introduction_section}

\begin{figure*}
\includegraphics[scale = 0.8]{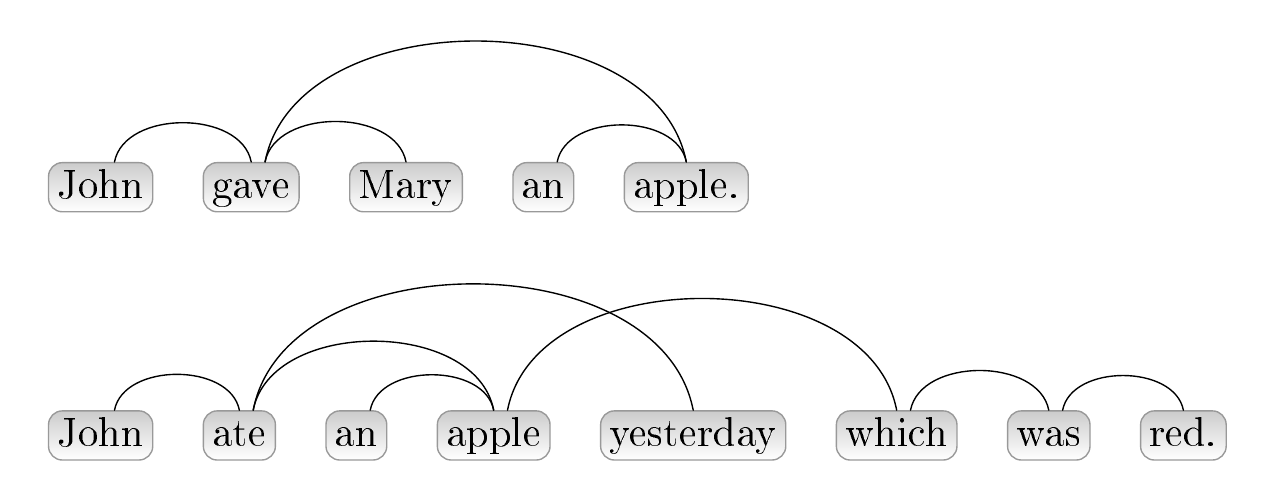}
\caption{\label{syntactic_dependency_trees_figure} Syntactic dependency trees of sentences. Top: a tree without dependency crossings. "John" is the subject of the verbal form "gave". Center: a tree with one edge crossing (the crossing is formed by the edge linking "ate" and "yesterday" and the edge linking "apple" and "which"). Bottom: the same sentence after fronting "yesterday". Notice that the length of the dependency between "ate" and "yesterday" and the dependency between "apple" and "which" have reduced (while the length of other dependencies has remained constant). Top and center are adapted from \cite{Ambati2008a}. }
\end{figure*}

The syntactic dependency structure of a sentence can be defined as a network where vertices are words and connections indicate syntactic dependencies, e.g., the relationship between the subject of a sentence and its verb (Fig. \ref{syntactic_dependency_trees_figure}). These networks are typically trees and directed \cite{tesniere59,hays64,melcuk88}. However, link direction is irrelevant for the present article and therefore omitted. Syntactic dependency networks can be seen as 
spanning trees on a lattice \cite{Manna1992a,Barthelemy2006a} and are indeed a particular case of geographically embedded or spatial networks \cite{Reuven2010a_Chapter8, Gastner2006a, Guillier2017a} 
in one dimension, i.e. the dimension defined by the linear order of the words of the corresponding sentence \cite{Ferrer2004b}.

In the context of syntactic dependency networks, the length of an edge is defined as an Euclidean distance, namely, 
the linear distance between the words that are connected: adjacent words are at distance 1, words separated by one word are at distance 2, and so on \cite{Ferrer2004b,Liu2017}. In the sentence at the top of Fig. \ref{syntactic_dependency_trees_figure}, {\em John} and {\em gave} are at distance 1 while {\em gave} and {\em apple} are at distance 3. 

\begin{figure}
\begin{center}
\includegraphics[scale = 0.8]{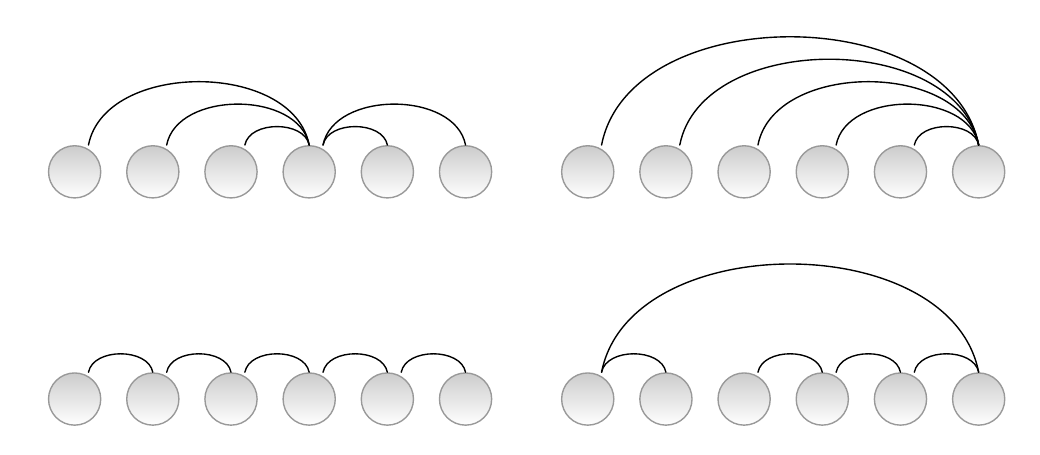}
\caption{\label{star_and_linear_trees_figure} Linear arrangements of trees with $n = 6$ vertices. Top: A star tree, a tree with a vertex of maximum degree \cite{Ferrer2013d}. Bottom: A linear tree, a tree where vertex degrees do not exceed 2 \cite{Ferrer2013d}. }
\end{center}
\end{figure}

Syntactic dependency trees exhibit certain statistical patterns concerning the length of their dependencies and the variance of their degrees. First, edge lengths are biased towards low values \cite{Ferrer2004b,Ferrer2003f} as it happens in other geographical networks \cite{Gastner2006a}. The distribution of edge lengths decays exponentially \cite{Ferrer2004b} as is the case of the distribution of projection lengths in real neural networks \cite{Ercsey2013a}. Additionally, the mean edge length is smaller than expected by chance \cite{Ferrer2004b, Ferrer2013c, Liu2008a, Futrell2015a, Liu2017}.  The simplest null hypothesis assumes a uniformly random permutation of the words of a sentence and predicts that the expected edge length is $(n + 1)/3$, where $n$ is the number of vertices of the tree (the length of the sentence in words) \cite{Ferrer2004b,Zornig1984a}.  
Second, their hubiness coefficient does not exceed $25\%$ \cite{Ferrer2017a}. $h$, the hubiness coefficient is a normalized variance of vertex degrees. $h$ is a number between 0 and 1 that is minimum for linear trees and maximum for star trees (Fig. \ref{star_and_linear_trees_figure}). 
Indeed, the hubiness of real syntactic dependencies is close to trees from the ensemble of uniformly random trees, for which $h$ tends to zero as $n$ increases \cite{Ferrer2017a}.

The target of the present article are the edge crossings that can arise when drawing connections above the sentence. Fig. \ref{syntactic_dependency_trees_figure} shows two planar sentences (a sentence is planar if it does not have crossings) and a sentence with one crossing. It is widely accepted that crossing dependencies are relatively uncommon in languages \cite{lecerf60,hays64,melcuk88,Ferrer2006d,Park2009a,Liu2010a,Gildea2010a}. 
Indeed, the actual number of crossings per sentence does not reach $3.5$ across languages and is only above $1$ in a few of them \cite{Ferrer2017a}. However, how small a number is depends on the scale of measurement and a null model is required. A rigorous demonstration that crossings are really scarce has been missing for decades. Recently, statistical evidence that crossings are significantly small has been provided \cite{Ferrer2017a}. Furthermore, sentences where dependencies are shorter tend to have fewer crossings \cite{Ferrer2015c}. Fig. \ref{syntactic_dependency_trees_figure} (center and bottom) illustrates the tendency of crossings to reduce as dependency lengths reduce.



Research on syntactic dependency networks parallels research on non-spatial networks: as the statistical properties of many real networks have been compared against the predictions of null models, the Erd\H{o}s-R\'enyi graph being one of the most simple and popular examples \cite{Newman2010a}, the statistical properties of syntactic dependency networks have been compared against the predictions of null models with increasing levels of complexity for the length of syntactic dependencies \cite{Ferrer2004b, Liu2008a, Futrell2015a} or for the number of crossings \cite{Ferrer2017a, Ferrer2014f, Ferrer2014c}.

Beyond network theory, the issue of the presence and frequency of crossing dependencies in the syntax of natural languages has received considerable attention in the computational linguistics community, as supporting them makes parsing computationally harder \cite{McDonald07,Hav07,GomCL2016}. Crossings are also relevant in biology, where they appear in networks of nucleotides whose vertices are occurrences of nucleotides $A$, $G$, $U$, and $C$ while edges are Watson-Crick ($A$-$U$, $G$-$C$) and $U$-$G$ base pairs \cite{Chen2009a}. 

In this context, a question naturally arises: what is the reason for the low frequency of crossing dependencies, consistently observed across languages? A traditional answer consists of postulating that there is some kind of grammatical ban on 
crossing dependencies \cite{sleator93,Tanaka97,KyotoCorpus,Starosta03,Lee04,hudson07,Ninio2017}.
However, this position fails to explain many linguistic phenomena involving crossings 
\cite{Versley2014,Levy2012a}. Another option is to assume that crossing dependencies can be grammatical, but only if they follow certain patterns or hard constraints. However, while some classes of non-crossing dependency structures have a very good empirical coverage of real sentences \cite{kuhlmann06,gomez2011cl,GomNiv2013,GomCL2016}, these proposals still face counterexamples that fall outside the restrictions \cite{chen2010unavoidable,Bhat2012,ChenMain2014}.

From the perspective of theoretical linguistics, the grammatical ban on crossings can be interpreted:
\begin{itemize}
\item
As a ban set independently from performance considerations, e.g., requiring some hidden parameter to be turned. In this case the ban can be seen as avoidable (e.g., it depends on whether the parameter is on or off for each given language). 
\item
As a consequence of performance constraints associated directly to crossing dependencies. The ban would be inevitable if the cognitive pressures were strong enough but then it would not be properly a ban (a norm added on top of human cognition) but rather a side-effect of cognitive constraints. This view is challenged by psychological and graph theoretic research indicating that crossing dependencies can be easier to process (\cite{deVries2012a} and \cite{Ferrer2014f} and references therein).
\end{itemize}  
Some researchers have adopted an apparently neutral position concerning the nature of the ban but assume that the low frequency of crossings derives from an independent and specific constraint on crossings: explicitly when postulating a principle of minimization of crossings \cite{Liu2008a} or implicitly in a large body of research on dependency length minimization that takes for granted that syntactic dependencies should not cross \cite{Liu2008a,Park2009a,GildeaTemperley10,Futrell2015a,Gulordava2015}. 

If it turned out that non-crossing dependencies can be explained as a side-effect of some cognitive pressure that is not directly associated to crossings (e.g., dependency length minimization), could all these views be regarded as really neutral regarding the nature of the ban?  

In this article, we explore a simpler hypothesis: that in order to explain the scarcity of crossing dependencies in language, it is not necessary to assume any underlying rule or principle of human languages that is responsible directly for this fact (including the possibility of some cognitive cost associated directly to crossings). 
Instead, the low frequency of crossings may naturally arise, indirectly, from the actual length of dependencies \cite{Ferrer2015c}, which are constrained by a well-known  psychological principle: dependency length minimization (see \cite{Liu2017}, \cite{Ferrer2013e} or \cite{Tily2010a} for a review). That explanatory principle, which holds even in languages allowing for words to scramble freely \cite{Futrell2015a}, could follow from more general constraints on language processing \cite{Christiansen2015a}. 

As dependency length minimization can be seen as particular case of minimization of the Euclidean distance between connected vertices in an $m$-dimensional space, our originally linguistic problem on crossings is related to the general problem of minimizing the cost of load transportation over a network in complex systems science \cite{Guillier2017a} and the minimum linear arrangement problem of computer science \cite{Ferrer2004b,Ferrer2016a}.

To investigate the origins of the scarcity of crossing dependencies, we use treebanks (collections of sentences with their corresponding syntactic dependency network) to provide statistical evidence that the amount of dependency crossings in a wide range of languages can be predicted with small error by a simple estimator based exclusively on dependency length information and information on which edges can potentially cross (edges that share a vertex cannot cross). 

We will show that the estimator consistently delivers good predictions of the number of crossings, 
in two different collections of dependency treebanks with diverse annotations. An annotation is a set of criteria used to define the syntactic dependency structure of a sentence. We will argue that this is the best explanation for the low frequency of crossings when both psychological plausibility and parsimony at all levels (from a model of crossings to a general theory of language) are required. Our predictor is a null model in the sense that for every pair of edges that may potentially cross it assumes that the corresponding vertices take random positions in the linear sequence of the sentence. 

The remainder of the article is organized as follows. Section \ref{predictors_section} discusses various ways in which the crossings of a sentence could be predicted. 
Section \ref{crossing_theory_section} presents the predictor of crossings chosen for this article and its theoretical background. The dependency trees used to test the predictor are presented in Section \ref{methods_section}. Section \ref{results_section} shows the results of the predictions, and Section \ref{discussion_section} discusses some implications for computational linguistics and linguistic theory.    

\section{Possible predictors}

\label{predictors_section}

Here we will examine various possibilities to predict the number of dependency crossings in a sentence. 
The problem of the origins of non-crossing dependencies can be recast as problem of modeling: we want to find the best model
for predicting the number of crossings in a sentence. According to modern model selection, the best model is the one that has the best trade-off between quality of fit (predictive power) and parsimony \cite{Burnham2002a}. 
We complement this view
involving further requirements: 
\begin{itemize}
\item
The model must be psychologically realistic. A model that assumes orderings of words that are hard to produce by the human brain should be penalized with respect to one that is based on orderings that real speakers produce (or can rather easily produce). We are not only simply concerned about predicting the low number of crossings of a sentence but also understanding why that number is that low. Hiding the problem under the carpet of grammar or an ad-hoc principle of planarity 
does not help.
\item  
Its assumptions must be valid. The predictions of a model may be compatible with real data and even be of high quality but its assumptions may not be supported by real data or inconsistent with the source that produced it.  
\item 
We are not only concerned about the best model in a local sense but one that leads to a general theory of word order or even a comprehensive theory of language that is compact. A real scientific theory is more than a collection of disconnected ideas or models \cite{Bunge2001a_French}.
Models that lead to an unnecessarily fat general theory when integrated into it should also be penalized. 
Models that exploit assumptions from successful models in other domains should be favored. 

For instance, a model that allows one to understand not only the scarcity of crossings but also why adjectives tend to be placed 
before the noun in SOV languages is preferable to one that requires an independent solution to explain the placement of adjectives \cite{Ferrer2014f}. SOV languages are languages that tend to put the subject (S) before the object (O) and in turn, O before the verb (V) under some general conditions \cite{wals-81}.
\end{itemize}

In what follows, we will use $C$ to refer to the number of dependency crossings in the parse of a sentence (i.e., the number of pairs of syntactic dependencies that cross). Our goal is, therefore, to find a suitable predictor for $C$. Note that $C = 0$ for a star tree \cite{Ferrer2013d}. The sum of the lengths of all dependencies in a sentence will be denoted by $D$. 

\subsection{Minimization of crossings}

A principle of minimization of crossings \cite{Liu2008a} leads to a simple deterministic predictor: $C = 0$, reflecting a grammatical ban on crossings 
\cite{sleator93,Tanaka97,KyotoCorpus,Starosta03,Lee04,hudson07}.

This predictor is problematic for various reasons:
\begin{itemize}
\item
Concerning the validity of its assumptions, the model assumes that $C = 0$ independently from $D$, while $C$ and $D$ are positively correlated in many languages \cite{Ferrer2015c}. 
\item
Concerning the accuracy of its predictions, this model fails because sentences with $C > 0$ are found in many languages \cite{Ferrer2017a} and the likelihood of the model is minimum, which indicates that the model is among the worst possible models for crossing dependencies according to modern model selection \cite{Burnham2002a} because its likelihood is zero. Furthermore, 
the model fails to explain many linguistic phenomena involving crossings 
\cite{Versley2014,Levy2012a}.  
\item
Its psychological validity is unclear. If the model is interpreted as arising from processing difficulties inherent to crossing dependencies \cite{Levy2012a} or computational tractability (as reviewed in Section \ref{introduction_section}) then it is challenged by psychological and graph theoretic research indicating that sentences with $C >0$ can be easier to process than sentences with $C= 0$ (see \cite{deVries2012a}, \cite{Ferrer2014f}, \cite{Chung1978a} and references therein). Another problem is how a language generation process could warrant that $C=0$. If $C=0$ is determined before the sentence is produced, how is it possible that sentence production does not introduce (many) crossings? Crossing theory indicates that a star tree is needed to keep a low number of crossings \cite{Ferrer2014f}. 
If $C=0$ is determined while the sentence is produced (linearized), how are crossings avoided on the fly as real language production is not a batch process \cite{Christiansen2015a}? It looks simpler to consider that non-crossing dependencies are a side-effect of a principle of dependency length minimization \cite{Ferrer2006d,Ferrer2014c,Ferrer2014f}.
\item
Concerning the compactness of the whole theory, the model $C = 0$ leads to a fatter theory of language because the scarcity of crossings and also the positive correlation between $D$ and $C$ could be explained to a large extent by recycling the highly predictive principle of dependency length minimization \cite{Ferrer2013e}, as we will see below. 
\end{itemize}

Another option is to assume that crossing dependencies can be grammatical, but only if they follow certain patterns or hard constraints. However, while some 
classes of dependency structures tolerating certain crossings
have a very good empirical coverage \cite{kuhlmann06,gomez2011cl,GomNiv2013,GomCL2016}, these proposals still face counterexamples that fall outside the restrictions \cite{chen2010unavoidable,Bhat2012,ChenMain2014}.

One possibility is to relax the simple deterministic predictor above so that on average $C = \gamma$, where $\gamma$ is a constant, e.g., $\gamma = 3.3$ as in  ancient Greek \cite{Ferrer2017a}. However, it has been shown that this is problematic because $C = \gamma$ might be impossible to reach if $n$ is sufficiently small (see Appendix of \cite{Ferrer2015c}). Therefore, a proper relaxation of this deterministic predictor is  $C = \gamma(n)$, where $\gamma$ is a function that only depends on $n$ \cite{Ferrer2015c}.
This allows one to capture the variation in the number of crossings across languages, but adding extra parameters, and it is still problematic for the reasons of the case $\gamma(n) = 0$ that we have examined above. Further arguments can be found in Section 4.3 of \cite{Ferrer2014f}. 

\subsection{Minimum linear arrangement}

A minimum linear arrangement of a sentence is an ordering of the words of the sentence that minimizes the sum of dependency lengths.
One may predict the assumed number of crossings by calculating the minimum linear arrangements of a sentence \cite{Ferrer2006d}. A possible predictor could be the mean number of crossings over all those arrangements. 

The predictive power of the model is supported by the fact that solving the minimum linear arrangement problem reduces crossings to practically zero  \cite{Ferrer2006d}, as in many languages. A potential problem of this model is that it has never been checked whether it predicts the actual number of crossings of real sentences, as far as we know.  

Perhaps the major challenge for this predictor is the validity of the assumption of a minimum linear arrangement because:
\begin{itemize}
\item
The actual value of $D$ in real sentences is located between the minimum and that of a random ordering of vertices \cite{Ferrer2004b,Ferrer2013c}. The ratio $\Gamma = D/D_{min}$ (where $D_{min}$ is the minimum value of $D$) is greater than 1.2 in Romanian for sufficiently long sentences \cite{Ferrer2004b} and a similar lower bound on language efficiency has been found in English across centuries \cite{Tily2010a}. 
\item
It may not be valid also for theoretical reasons: word order is a multiconstraint satisfaction problem where the principle of dependency length minimization is in conflict with  other word order constraints \cite{Ferrer2014a,Futrell2015a}.
Thus, a model based on minimum linear arrangements is not that simple: it has to explain why dependency length minimization dominates fully over other principles or provide evidence that the distortion caused by other principles can be neglected. Below we will present a model that does not have this problem because it works on true dependency lengths, which are expected to be determined by the interplay between dependency length minimization and other principles.  
\item
The full minimization of $D$ is cognitively unrealistic, as it is incompatible with the predictions of the now-or-never bottleneck \cite{Christiansen2015a}. As for the latter, notice that the minimization of $D$ implies that the whole sentence must be available as input for some minimum linear arrangement algorithm, whereas actual language generation and processing is intrinsically online and heavily constrained by our fleeting memory \cite{Christiansen2015a}.
\end{itemize}

\subsection{Random linear arrangement}

If the minimum linear arrangement is too restrictive, one could consider the opposite:  
predicting the number of crossings assuming a random ordering of the words of the sentence \cite{Ferrer2014f}. However, a random linear arrangement cannot explain the low numbers of crossings observed in real sentences. Empirically, the number of crossings of sentences is much smaller than the number of crossings expected by random linear arrangement \cite{Ferrer2017a}. Theoretically, a constant low number of crossings requires a star tree \cite{Ferrer2014f}. 

The failure of a random linear arrangement is not surprising. First, it 
is cognitively unrealistic: even in languages with high word order flexibility, word order is constrained \cite{wals-81,Lester2015a}. Second, the assumption that the ordering of sentences is arbitrary (unconstrained) is easily rejected by the fact that dependency lengths are below chance in real languages \cite{Ferrer2004b,Ferrer2013c,Futrell2015a}. 
Thus, this predictor is only useful as a random baseline for other predictors. Here we will compare it against a better predictor that is introduced next. 
  
\subsection{Random linear arrangement with some knowledge about dependency lengths}

A stronger predictor can be built by focusing on the set of pairs of edges that may potentially cross and basing predictions on the actual length of the edges under the assumption of a random linear arrangement of the four vertices that are potentially involved in an edge crossing \cite{Ferrer2014c}. So far, its predictive power is supported by its capacity to predict the actual number of crossings in random trees with an error of about $5\%$ \cite{Ferrer2014c}.
A crucial goal of the present article is to test the accuracy of its predictions on real sentences. 
This predictor is promising because actual dependency lengths are below chance, i.e. below $(n+1)/3$ \cite{Ferrer2004b,Ferrer2013c}, a domain where the probability theory behind this model indicates that shortening a dependency yields a reduction in the probability that it crosses a dependency of unknown length in a random linear arrangement of the two edges (Section 5 of \cite{Ferrer2014f}).

For the reason above,
this predictor is fully compatible with the positive correlation between $D$ and $C$ \cite{Ferrer2015c,Ferrer2014f}, in contrast with the deterministic predictor ($C=0$) and its generalization. 
Concerning assumptions, this model is simpler than the model based on minimum linear arrangements: this model does not assume an unrealistic ordering of the elements of the sentence but the true ordering.  
Its psychological validity is greater than that of the minimum linear arrangement predictor because it can base its prediction on information from sentences that have actually been produced by a speaker or a writer. 
Contrary to the minimum linear arrangement predictor, this model bases its prediction on true dependency lengths instead of ideal values of $D$.

However, it can be argued that a fundamental assumption of the model, namely that vertices are arranged linearly at random, is not supported empirically, following the arguments against the random linear arrangement predictor. This is a fair criticism, but for this reason this model should be regarded as a null hypothesis rather than as a fully realistic model. 

Having said that,
modeling requires a compromise between quality of fit, adequacy and parsimony. If this null hypothesis model provides predictions of sufficient quality on real sentences, do we really need to worry about providing a more realistic but also more complicated model? 
Put differently, suppose that the information provided by the lengths of edges suffices to predict reasonably well the low number of crossings of real sentences, even without assuming any additional constraint on the linear arrangement of the involved vertices that could help to minimize crossings, but instead modeling it under the weakest possible assumption (namely placing vertices at random). Then considering more fine-grained information or more realistic orderings is secondary to our particular goal. In the worst case, this predictor would be an inevitable baseline for an alternative model. 

Before we proceed, it is worth noting that our article is not a mere application of an established model or theory to a concrete dataset, but the first test of a novel theory on a massive collection of networks from different languages and different annotation criteria, which has implications to our understanding of the faculty of language as such.  
The result of such a test is far from trivial, and thus its success is a relevant contribution, for two reasons.
First, our model, which is a null model rather than a realistic model, assumes that vertices are arranged purely at random in a sequence (preserving the original edge lengths). However, real sentences are not random sequences of words, as research on long correlations in physics has been showing for more than a decade, e.g. \cite{Montemurro2001b,Altmann2012a}. 
Second, although such null model predictor has been tested previously on uniformly random trees \cite{Ferrer2014c}, one cannot assume the predictor will work on real sentences given the substantial statistical differences between uniformly random trees and real syntactic dependency trees \cite{Ferrer2017a}.

The next section introduces the mathematical definition of the promising predictor above and its theoretical background. 

\section{Crossing theory}

\label{crossing_theory_section}

Here we provide a quick overview of a crossing theory developed in a series of articles \cite{Ferrer2013b,Ferrer2013d,Ferrer2014c,Ferrer2014f, Ferrer2017a}. 
It is correct to state that $C$ cannot exceed the number of pairs of different edges, namely
\begin{equation}
C \leq {n - 1 \choose 2}
\label{naive_maximum_number_of_crossings_equation}
\end{equation}  
However, the truth is that 
\begin{equation}
C \leq {n - 2 \choose 2}
\label{maximum_number_of_crossings_equation}
\end{equation}
with equality in case of a linear tree (see \cite{Ferrer2017a} for linear arrangements of linear tree that maximize $C$). The upper bound above is defined based only on knowledge of the size of the tree. Adding further properties of the tree, the upper bound can be refined.

A central concept of crossing theory is $Q$, the set of pairs of edges of a tree that can potentially cross when their vertices are arranged linearly in some arbitrary order (edges sharing a vertex cannot cross).  
$|Q|$, the cardinality of $Q$, is the potential number of crossings, i.e. 
\begin{equation}
C \leq |Q| \leq {n - 2 \choose 2}.
\end{equation}
We have  
\begin{equation}
|Q| = \frac{n}{2} \left(\left< k^2 \right>_{star} - \left< k^2 \right>\right),
\label{potential_number_of_crossings_equation}
\end{equation}
where 
$\left< k^2 \right>$ is the mean of the squared degrees of its vertices and $\left< k^2 \right>_{star}= n - 1$ is the value of $\left< k^2 \right>$ in a star tree of size $n$ \cite{Ferrer2014c,Ferrer2013d}. $|Q| = 0$ if and only if the tree is a star tree \cite{Ferrer2013d}.

With the theoretical background above, it is easy to see why $C$ cannot exceed the number of different pairs that can be formed out of $n-2$ elements (Eq. \ref{maximum_number_of_crossings_equation}) instead of $n-1$, that coincides with the number of edges (Eq. \ref{naive_maximum_number_of_crossings_equation}): that is the conclusion of computing the value of $\left< k^2 \right>$ for a linear tree, i.e. $\left< k^2 \right>_{linear} = 4 - 6/n$, and then applying 
$\left< k^2 \right> \geq 4 - 6/n$ to Eq. \ref{potential_number_of_crossings_equation} 
\cite{Ferrer2017a}.

$C$ denotes the number of crossings of the linear arrangement of a graph in general while  
$C_{true}$ denotes the number of crossings of the syntactic dependencies of a real sentence. The relative number of crossings is $\bar{C} = C/|Q|$ or 
$\bar{C}_{true} = C_{true}/|Q|$ \cite{Ferrer2014c}. $C$ can be expressed as a sum over $Q$, i.e. 
\begin{equation}
C = \sum_{(e_1, e_2) \in Q} C(e_1, e_2),
\label{number_of_crossings_equation}
\end{equation} 
where $C(e_1, e_2)$ is an indicator variable, $C(e_1, e_2)$ = 1 if the edges $e_1$ and $e_2$ cross and $C(e_1, e_2) = 0$ otherwise.  
The simplest prediction about $C$ than can be made departs from the null hypothesis that the vertices are arranged linearly at random (all possible orderings are equally likely). Following Eq. \ref{number_of_crossings_equation}, the expected number of crossings under that null hypothesis turns out to be 
\begin{eqnarray}
E_0[C] & = & \sum_{(e_1, e_2) \in Q} E[C(e_1, e_2)]\\
       & = & \sum_{(e_1, e_2) \in Q} p(C(e_1, e_2) = 1), 
\label{expected_number_of_crossings_equation}
\end{eqnarray}
where $p(C(e_1, e_2) = 1)$ is the probability that the edges $e_1$ and $e_2$ cross knowing that they belong to $Q$. 
Under that null hypothesis, the probability that two edges of $Q$ cross is constant, i.e. $p(C(e_1, e_2) = 1) = 1/3$, yielding \cite{Ferrer2013d} $E_0[C] = |Q|/3$.

\begin{figure*}
\includegraphics[scale = 0.8]{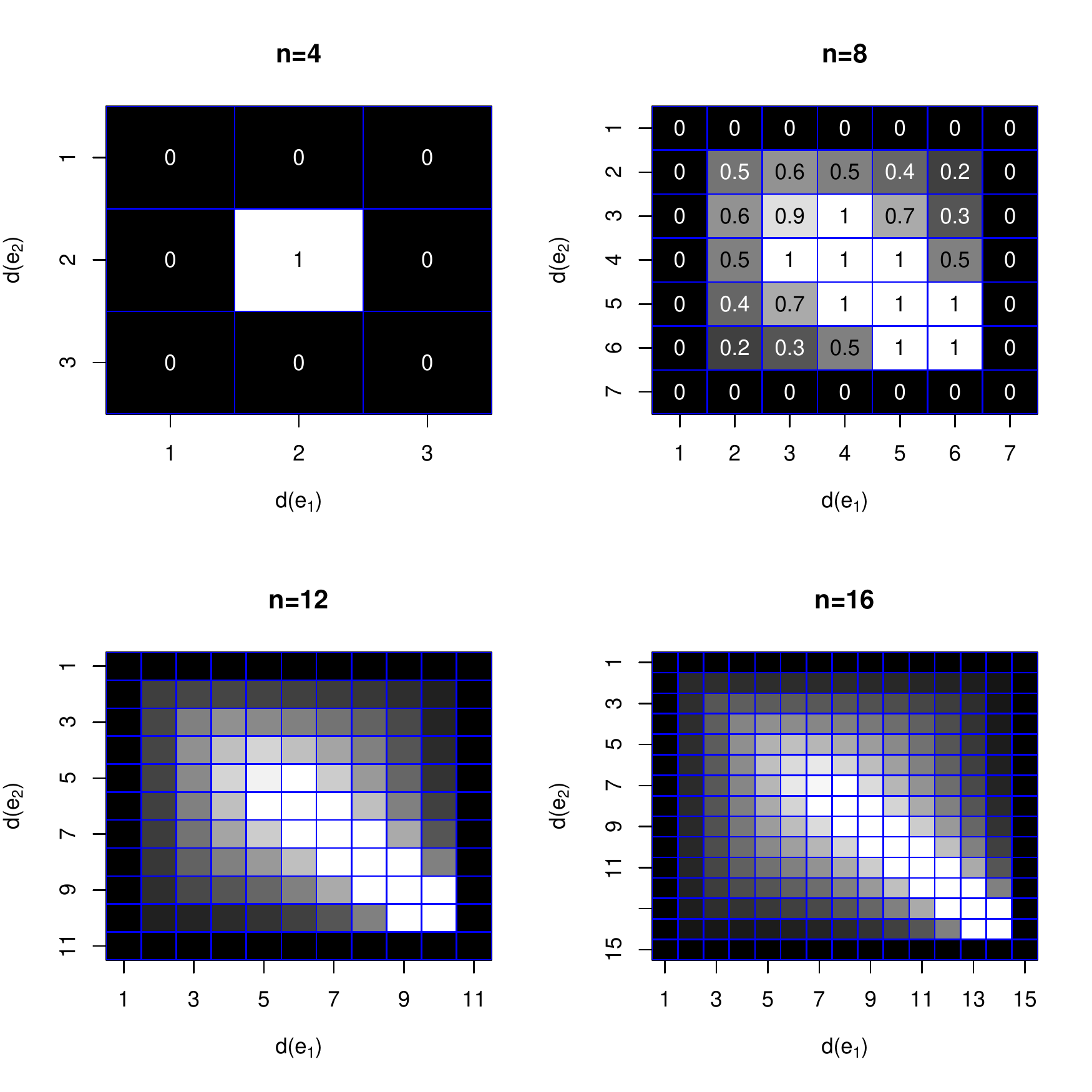}
\caption{\label{probability_given_two_lengths_figure} $p(C(e_1, e_2) = 1 | d(e_1), d(e_2))$
as a function of $d(e_1)$ and $d(e_2)$ for different values of $n$ (tree size in vertices). 
$d(e)$ is the length of the edge $e$ and $p(C(e_1, e_2) = 1 | d(e_1), d(e_2))$ is the probability that $e_1$ and $e_2$ (a pair of edges of $Q$) cross in a random linear arrangement of their vertices, knowing their lengths. Brightness is proportional to $p(C(e_1, e_2) = 1 | d(e_1), d(e_2))$ (black for $p(C(e_1, e_2) = 1 | d(e_1), d(e_2)) = 0$ and white for $p(C(e_1, e_2) = 1 | d(e_1), d(e_2)) = 1$). The two panels on top show the value of the probability.} 
\end{figure*}

The prediction offered by $E_0[C]$ can be improved by introducing knowledge about the length of the dependencies (e.g., edges of length 1 or $n - 1$ are not crossable). Suppose that $d(e)$ is the length of the edge $e$ and that $p(C(e_1, e_2) = 1 | d(e_1), d(e_2))$ is the probability that $e_1$ and $e_2$ (two arbitrary edges of $Q$) cross in a random linear arrangement of their vertices knowing their lengths.
The predictor $E_2[C]$ is obtained when $p(C(e_1, e_2) = 1)$ is replaced by $p(C(e_1, e_2) = 1 | d(e_1), d(e_2))$
in Eq. \ref{expected_number_of_crossings_equation}, yielding 
\begin{equation}
E_2[C] = \sum_{(e_1, e_2) \in Q} p(C(e_1, e_2) = 1 | d(e_1), d(e_2)),
\label{predicted_number_of_crossings_equation}
\end{equation}
$p(C(e_1, e_2) = 1 | d(e_1), d(e_2))$ depends only on $n$, $d(e_1)$ and $d(e_2)$ and
is defined as  
\begin{equation}
p(C(e_1, e_2) = 1 | d(e_1), d(e_2)) = \frac{|\alpha(d(e_1),d(e_2))|}{|\beta(d(e_1),d(e_2))|},
\label{probability_of_crossing_given_two_lengths_equation}
\end{equation}
where here $|..|$ is the cardinality operator, $\alpha(d_1,d_2)$ is the set of valid pairs of initial positions of two edges of lengths $d_1$ and $d_2$ that involve a crossing and $\beta(d_1,d_2)$ is the set of valid pairs of initial positions of edges of lengths $d_1$ and $d_2$, thus $\alpha(d_1,d_2) \subseteq \beta(d_1,d_2)$. Fig. \ref{probability_given_two_lengths_figure} shows a two-dimensional map of $p(C(e_1, e_2) = 1 | d(e_1), d(e_2))$. The perimeter of the map contains zeroes because an edge of minimum length ($1$) or maximum length ($n-1$) cannot cross any other edge. The map is symmetric with respect to the diagonal that crosses the top-left corner and the bottom-right corner by symmetry, namely
\begin{eqnarray}
p(C(e_1, e_2) = 1 | d(e_1), d(e_2)) = \notag \\  
                                    p(C(e_2, e_1) = 1 | d(e_2), d(e_1)). 
\end{eqnarray}
The map for $n = 4$, the minimum value of $n$ needed to have $|Q| > 0$, shows that only edges of length 2 can cross. The maps for $n = 8$, $n = 12$ and $n=16$ show that a reduction of the length of one of the edges causes the probability of crossing to reduce if edge lengths are sufficiently small. This reduction of the probability of crossings is likely to occur in real languages, where the mean length of dependencies is on average smaller than the random baseline $(n+1)/3$ \cite{Ferrer2004b, Ferrer2013c} and long edges would imply a cognitive cost that may not be afforded \cite{Liu2017,Christiansen2015a,Ferrer2013e}.

Although $E_0[C]$ and $E_2[C]$ are predictors of $C$ that have the same mathematical structure (they are sums of probabilities over pairs of edges of $Q$), $E_0[C]$ is a true expectation while $E_2[C]$ is not. 

The relative error of a predictor is defined as \cite{Ferrer2014c}
\begin{equation}
\Delta_x = E_x\left[\bar{C}\right] - \bar{C}_{true} = \frac{1}{|Q|}\left(E_x[C]-C_{true}\right).
\label{error_equation}
\end{equation}
$\Delta_0$ will be used as a baseline for $\Delta_2$. Interestingly,
$\Delta_0$ converges to $1/3$ for sufficiently long sentences when $C_{true}$ is bounded by a constant and $|Q|$ is large enough. The reason is that $E_0[C]/|Q| = 1/3$ and then 
\begin{equation}
\Delta_0 = \frac{1}{3} - C_{true}/|Q|.
\end{equation}
That explains why $\Delta_0$ converges to $1/3$ for sufficiently large $n$ in uniformly random trees where $C$ is bounded by a small constant \cite{Ferrer2014c} because uniformly random trees have a high $|Q|$, or equivalently, a low hubiness \cite{Ferrer2017a}. We also expect $\Delta_0$ to converge to $1/3$ in real syntactic dependency trees because $C_{true}$ is small and their hubiness is also low \cite{Ferrer2017a}. 

\section{Materials and methods}

\label{methods_section}

We considered the corpora in version 2.0 of the HamleDT collection of treebanks \cite{HamleDTJournal,HamledTStanford}. This collection is a harmonization of existing treebanks for 30 different languages into two well-known annotation styles: Prague dependencies \cite{PDT20} and Universal Stanford dependencies \cite{UniversalStanford}. Therefore, this collection allows us to evaluate our predictions of crossings both across a wide range of languages and two popular annotation schemes. The latter is useful because observations like the number of dependency crossings present in treebank sentences do not only depend on the properties of languages themselves, but also on annotation criteria (\cite{Ferrer2015c} lists some examples of how annotation criteria may affect $C$).

Each of the syntactic dependency structures in the treebanks was preprocessed by removing nodes corresponding to punctuation tokens, as it is standard in research related to dependency length (e.g., \cite{Ferrer2004b,Ferrer2015c,Futrell2015a}), which is only concerned with dependencies between actual words. To preserve the syntactic structure of the rest of the nodes, non-punctuation nodes that had a punctuation node as their head were attached as dependents of their nearest non-punctuation ancestor. Null elements, which appear in the Bengali, Hindi and Telugu corpora, were also subject to the same treatment as punctuation.

After this preprocessing, a syntactic dependency structure was included in our analyses if (1) it defined a tree and (2) the tree was not a star tree. 
The reason for (1) is that our theory (e.g., $Q$) assumes a tree structure \cite{Ferrer2013d,Ferrer2014c} and that we wanted to avoid the statistical
problem of mixing trees with other kinds of graphs, e.g., the potential number
of crossings depends on the number of edges \cite{Ferrer2013b,Ferrer2014f,Ferrer2013d}. 
The reason for (2) is that crossings are impossible in a star tree \cite{Ferrer2013b}. Condition (2) implies that the syntactic dependency structure has at least four vertices (otherwise all the possible trees are star trees). By excluding star trees we are discarding trees where the prediction cannot fail.  
An additional reason for excluding star trees is that their relative number of crossings, $C/|Q|$, is not defined because
$C = |Q| = 0$.

Table \ref{relative_table} shows the number of sentences in the original treebanks and the number of sentences actually included in our analyses, after filtering by the criteria (1) and (2) above. The average number of crossings per sentence does not exceed 1 for most of the treebanks. See \cite{Ferrer2017a} for further details on the statistical properties of crossings in our collections of dependency treebanks. 

Here we adopt the convention of sorting languages in tables not alphabetically but decreasingly by number of crossings, measured according to the average number of crossings (the average $C_{true}$) with Stanford dependencies. It can be observed that languages that are known for their word order freedom, e.g., Latin or Ancient Greek, stand out on top of Table \ref{relative_table}. On the other hand, agglutinating languages like Basque, Japanese, Turkish, the Uralic languages Estonian and Finnish, and the Dravidian languages Tamil and Telugu, are placed rather to the bottom of the table. 
Agglutinating languages are languages where certain information is often integrated into words as morphemes (not leading to new vertices in the tree, except in the Turkish treebank) while non-agglutinating languages would instead place it in separate words (leading to separate vertices). Therefore, one expects fewer chances for dependency crossings in agglutinating languages, as equivalent information is expressed with fewer vertices, and the number of crossings tends to increase with the length of the sentence \cite{Ferrer2017a}. 

Our ordering by crossings should be taken as an approximation. For the sake of space, we only employ an ordering by crossings based on Stanford dependencies. Furthermore, the potential number of crossings may depend on factors such as genre, topic, sentence length or treebank size (number of sentences) and other biases \cite{Ferrer2015c,Ferrer2017a}. The collection of treebanks is heterogeneous in this respect. Therefore, the fact that one treebank has more crossings than another does not imply that the language of the former exhibits higher word order freedom than that of the latter. Other variables should be controlled for a more accurate ordering. Therefore, the focus of our article is on the power of the predictors in spite of the heterogeneity of the treebank collection. Linguistic distinctions such as agglutinating versus non-agglutinating languages are made to illustrate the potential of future linguistic research.
 
\section{Results}

\label{results_section}

\begin{figure*}
\includegraphics[scale = 0.8]{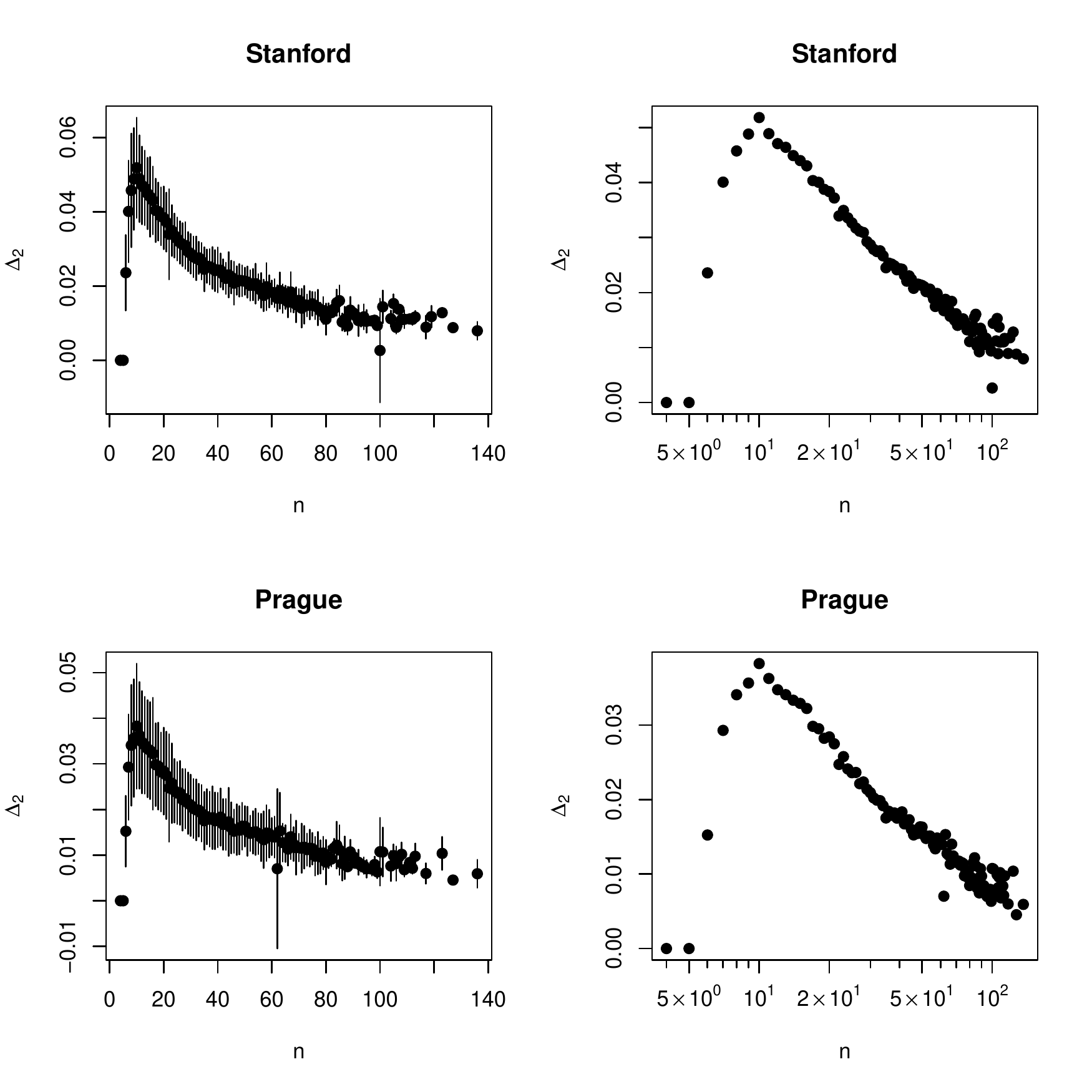}
\caption{\label{relative2_figure} $\Delta_2$, the error of the predictor based on the probability that two edges cross in a random linear arrangement preserving their original length, as a function of $n$, the size of the tree. Points and error bars indicate, respectively, mean values and $\pm 1$ standard deviation over proportions in a collection of treebanks. Tree sizes represented by less than two treebanks are excluded. Top: Stanford annotations. Bottom: Prague annotations.}
\end{figure*}

\begin{figure*}
\includegraphics[scale = 0.8]{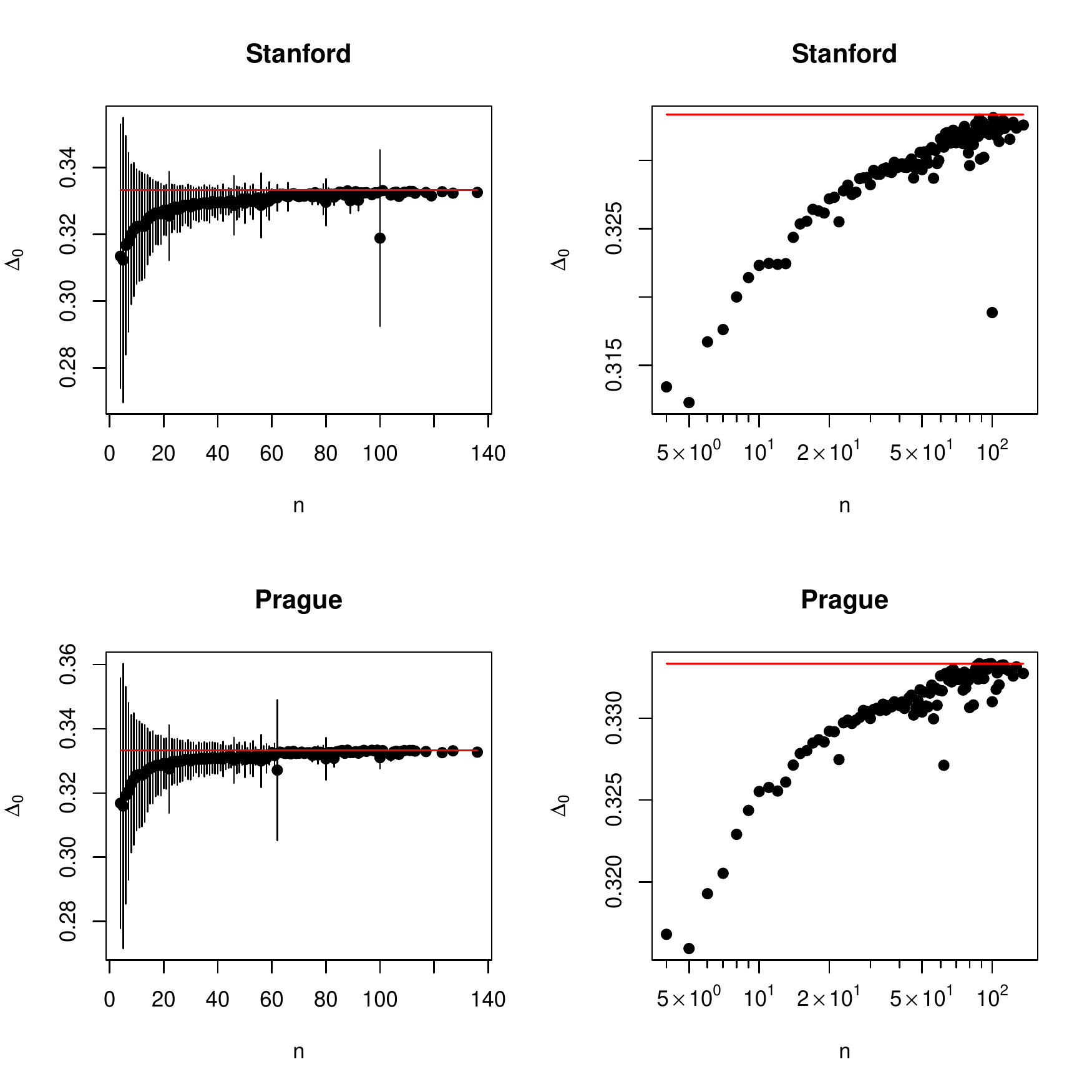}
\caption{\label{relative0_figure} $\Delta_0$, the error of the random linear arrangement predictor, as a function of $n$, the size of the tree. The format is the same as in Fig. \ref{relative2_figure}. A control line has been added to indicate 1/3, namely, the $\Delta_0$ that is expected when $C_{true}$ is bounded above by a small constant and $n$ goes to infinity. }
\end{figure*}

Figure \ref{relative2_figure} shows that, on average across treebanks, $\Delta_2$ increases as $n$ increases till $n = 10$ and decreases from that point onwards in both annotations. The maximum average $\Delta_2$ that is reached at $n = 10$ is $0.052$ for Stanford annotations and $0.038$ for Prague annotations. 
The predictor never fails for $n = 4$ and $n = 5$ ($\Delta_2 = 0$ in both cases) and from $n = 6$ onwards it always overestimates (on average) the actual number of crossings (recall Eq. \ref{error_equation}). Figure \ref{relative0_figure} shows that, on average across treebanks, $\Delta_0$ converges to $1/3$ as expected. 

Table \ref{relative_table} shows that the average $\Delta_2$, the relative error of the predictor $E_2[C]$,
is small: it does not exceed $5\%$. Thus, the average $\Delta_2$ is at least 6 times smaller than the baseline $\Delta_0 \approx 33\%$. 
The averages presented in Table \ref{relative_table} have been produced mixing measurements from sentences of different lengths. This is potentially problematic because the results might be heavily determined by the distribution of sentence lengths \cite{Ferrer2013c}. 

To control for sentence length, sentences were grouped by length and the average $\Delta_2$ was computed for the sentences within each group. Table \ref{relative_grouping_by_sentence_length_table} summarizes the statistical properties over the average $\Delta_2$ of each group. Interestingly, the average over group averages of $\Delta_2$ decreases with respect to the previous analysis: it does not exceed $4.3\%$. Thus, the average $\Delta_2$ is at least 7 times smaller than the baseline error, again $\Delta_0 \approx 33\%$. The minimum size of a group is one sentence; the qualitative results are very similar if the minimum size is set to 2.
      
\newcommand{\specialcell}[2][c]{%
  \begin{tabular}[#1]{@{}c@{}}#2\end{tabular}}

\begin{turnpage}
\begin{table*}
\caption{\label{relative_table} Summary of results for each treebank: number of sentences before and after filtering, average values of $C_{true}$ and $\Delta_0$, and average, median and standard deviation of the relative error $\Delta_2$, over the trees of each treebank. 
Romanian (Prague) is the only treebank with no crossing dependencies. Languages are sorted decreasingly by average $C_{true}$ according to Stanford dependencies.}

\begin{small}
\begin{center}
\begin{ruledtabular}
\begin{tabular}{lr|rccccc|rccccc} 
\phantom{x} & \phantom{x} & \multicolumn{2}{l}{\emph{Stanford annotation}} & \phantom{x} & \phantom{x} & \phantom{x} & \phantom{x} & \multicolumn{2}{l}{\emph{Prague annotation}} & \phantom{x} & \phantom{x} & \phantom{x} & \phantom{x} \\ \hline
Treebank & \#Sent & \specialcell{\#Sent\\ \footnotesize{(filtered)}} &
\specialcell{$C_{true}$\\ \footnotesize{(avg.)}} & \specialcell{$\Delta_0$\\ \footnotesize{(avg.)}} & \specialcell{$\Delta_2$\\ \footnotesize{(avg.)}} & \specialcell{$\Delta_2$\\ \footnotesize{(median)}} & \specialcell{$\Delta_2$\\ \footnotesize{(st. dev.)}} 
& \specialcell{\#Sent\\ \footnotesize{(filtered)}} &
\specialcell{$C_{true}$\\ \footnotesize{(avg.)}} & \specialcell{$\Delta_0$\\ \footnotesize{(avg.)}} & \specialcell{$\Delta_2$\\ \footnotesize{(avg.)}} & \specialcell{$\Delta_2$\\ \footnotesize{(median)}} & \specialcell{$\Delta_2$\\ \footnotesize{(st. dev.)}}\\ \hline
Anc. Greek & 21173 & 18713 & 3.2621 & 0.244 & 0.030 & 0.027 & 0.058  & 16237 & 3.3528 & 0.243 & 0.025 & 0.020 & 0.058 \\
Latin & 3473 & 3036 & 2.1785 & 0.282 & 0.034 & 0.031 & 0.046         & 2833 & 1.8503 & 0.286 & 0.036 & 0.032 & 0.047 \\
Dutch & 13735 & 10974 & 1.3980 & 0.311 & 0.046 & 0.041 & 0.051       & 11131 & 0.9898 & 0.315 & 0.034 & 0.027 & 0.041 \\
Hungarian & 6424 & 6103 & 0.9720 & 0.326 & 0.036 & 0.031 & 0.033     & 5047 & 0.8675 & 0.326 & 0.034 & 0.030 & 0.032 \\
Arabic & 7547 & 2280 & 0.9807 & 0.328 & 0.019 & 0.016 & 0.021        & 2248 & 0.0881 & 0.333 & 0.013 & 0.010 & 0.016 \\
German & 38020 & 33492 & 0.7826 & 0.325 & 0.050 & 0.046 & 0.036      & 32443 & 0.7230 & 0.326 & 0.043 & 0.039 & 0.033 \\   
Slovenian & 1936 & 1719 & 0.7749 & 0.322 & 0.047 & 0.039 & 0.046     & 1581 & 0.3125 & 0.327 & 0.035 & 0.027 & 0.038 \\     
Danish & 5512 & 4894 & 0.6800 & 0.324 & 0.047 & 0.040 & 0.038        & 4840 & 0.1643 & 0.331 & 0.027 & 0.022 & 0.027 \\        
Greek & 2902 & 2584 & 0.6540 & 0.330 & 0.039 & 0.033 & 0.028         & 2543 & 0.2057 & 0.332 & 0.030 & 0.024 & 0.023 \\    
Catalan & 14924 & 14520 & 0.6419 & 0.331 & 0.034 & 0.029 & 0.023     & 14556 & 0.0873 & 0.333 & 0.020 & 0.017 & 0.016 \\
Portuguese & 9359 & 8621 & 0.6336 & 0.328 & 0.039 & 0.033 & 0.032    & 8596 & 0.2465 & 0.331 & 0.021 & 0.016 & 0.021 \\
Spanish & 15984 & 15354 & 0.6218 & 0.331 & 0.034 & 0.029 & 0.024     & 15424 & 0.1105 & 0.333 & 0.020 & 0.016 & 0.017 \\
Persian & 12455 & 11579 & 0.5914 & 0.326 & 0.027 & 0.023 & 0.031     & 11632 & 0.4024 & 0.329 & 0.030 & 0.024 & 0.033 \\      
Czech & 87913 & 74843 & 0.5277 & 0.326 & 0.040 & 0.035 & 0.035       & 70023 & 0.3729 & 0.327 & 0.031 & 0.025 & 0.031 \\  
English & 18791 & 18275 & 0.5241 & 0.330 & 0.049 & 0.043 & 0.031     & 18369 & 0.1072 & 0.333 & 0.034 & 0.029 & 0.024 \\
Swedish & 11431 & 10714 & 0.4871 & 0.328 & 0.043 & 0.039 & 0.034     & 10207 & 0.1946 & 0.332 & 0.034 & 0.029 & 0.029 \\   
Slovak & 57408 & 47727 & 0.4559 & 0.324 & 0.044 & 0.036 & 0.044      & 44297 & 0.2688 & 0.326 & 0.034 & 0.026 & 0.039 \\     
Russian & 34895 & 31581 & 0.4171 & 0.326 & 0.038 & 0.032 & 0.035     & 31900 & 0.1570 & 0.330 & 0.027 & 0.021 & 0.028 \\   
Italian & 3359 & 2502 & 0.4153 & 0.329 & 0.035 & 0.029 & 0.032       & 2398 & 0.0621 & 0.333 & 0.020 & 0.014 & 0.024 \\     
Bulgarian & 13221 & 12119 & 0.3598 & 0.326 & 0.045 & 0.039 & 0.042   & 11947 & 0.1248 & 0.329 & 0.023 & 0.017 & 0.029 \\   
Finnish & 4307 & 4078 & 0.3183 & 0.326 & 0.034 & 0.028 & 0.038       & 4011 & 0.1279 & 0.330 & 0.028 & 0.024 & 0.031 \\        
Hindi & 13274 & 12417 & 0.3043 & 0.332 & 0.027 & 0.025 & 0.017       & 12334 & 0.3875 & 0.330 & 0.015 & 0.012 & 0.015 \\   
Japanese & 17753 & 4614 & 0.1641 & 0.326 & 0.024 & 0.019 & 0.032     & 4792 & 0.0002 & 0.333 & 0.006 & 0.000 & 0.013 \\    
Basque & 11225 & 9072 & 0.1391 & 0.330 & 0.028 & 0.022 & 0.033       & 8717 & 0.1252 & 0.330 & 0.026 & 0.021 & 0.029 \\    
Romanian & 4042 & 3145 & 0.1021 & 0.331 & 0.028 & 0.021 & 0.036      & 3193 & 0.0000 & 0.333 & 0.015 & 0.005 & 0.026 \\     
Bengali & 1129 & 678 & 0.1062 & 0.321 & 0.027 & 0.000 & 0.051        & 651 & 0.1244 & 0.320 & 0.025 & 0.000 & 0.052 \\      
Turkish & 5935 & 3862 & 0.0984 & 0.330 & 0.031 & 0.025 & 0.038       & 3518 & 0.1373 & 0.327 & 0.015 & 0.000 & 0.026 \\
Estonian & 1315 & 851 & 0.0376 & 0.331 & 0.016 & 0.000 & 0.037       & 843 & 0.0130 & 0.332 & 0.013 & 0.000 & 0.031 \\      
Tamil & 600 & 584 & 0.0240 & 0.333 & 0.026 & 0.022 & 0.025           & 585 & 0.0137 & 0.333 & 0.023 & 0.019 & 0.023 \\           
Telugu & 1450 & 429 & 0.0140 & 0.322 & 0.016 & 0.000 & 0.045         & 373 & 0.0080 & 0.325 & 0.014 & 0.000 & 0.043 \\      

\end{tabular}

\end{ruledtabular}
\end{center}
\end{small}
\end{table*}
\end{turnpage}

\begin{table*}
\caption{\label{relative_grouping_by_sentence_length_table} 
Summary of results for each treebank: number of distinct sentence lengths, 
average $\Delta_0$, and average, median and standard deviation of the average values of $\Delta_2$ over the groups of sentences with the same length. Languages are sorted decreasingly by average 
$C_{true}$ according to Stanford dependencies as in Table \ref{relative_table}.}
\begin{scriptsize}
\begin{center}
\begin{ruledtabular}
\begin{tabular}{l|rcccc|rcccc} 
\phantom{x} & \multicolumn{2}{l}{\emph{Stanford annotation}} & \phantom{x} & \phantom{x} & \phantom{x} & \multicolumn{2}{l}{\emph{Prague annotation}} & \phantom{x} & \phantom{x} & \phantom{x} \\ \hline
Treebank & \#Lengths & \specialcell{$\Delta_0$ \\ \footnotesize{(avg.)}} & \specialcell{$\Delta_2$ \\ \footnotesize{(avg.)}} & \specialcell{$\Delta_2$ \\ \footnotesize{(median)}} & \specialcell{$\Delta_2$ \\ \footnotesize{(st. dev.)}} 
& \#Lengths & \specialcell{$\Delta_0$ \\ \footnotesize{(avg.)}} & \specialcell{$\Delta_2$ \\ \footnotesize{(avg.)}} & \specialcell{$\Delta_2$ \\ \footnotesize{(median)}} & \specialcell{$\Delta_2$ \\ \footnotesize{(st. dev.)}} \\ \hline

Anc. Greek & 66 & 0.293 & 0.025 & 0.024 & 0.019  & 65 & 0.292 & 0.021 & 0.021 & 0.020 \\   
Latin & 59 & 0.309 & 0.031 & 0.029 & 0.018       & 59 & 0.313 & 0.031 & 0.030 & 0.017 \\        
Dutch & 54 & 0.319 & 0.037 & 0.035 & 0.019       & 54 & 0.323 & 0.027 & 0.024 & 0.016 \\        
Hungarian & 65 & 0.328 & 0.027 & 0.025 & 0.014   & 65 & 0.329 & 0.026 & 0.023 & 0.013 \\    
Arabic & 109 & 0.331 & 0.014 & 0.013 & 0.006     & 109 & 0.333 & 0.010 & 0.008 & 0.005 \\      
German & 85 & 0.328 & 0.033 & 0.032 & 0.012      & 85 & 0.329 & 0.029 & 0.027 & 0.011 \\       
Slovenian & 57 & 0.326 & 0.034 & 0.032 & 0.016   & 50 & 0.329 & 0.027 & 0.024 & 0.014 \\    
Danish & 66 & 0.328 & 0.031 & 0.029 & 0.013      & 66 & 0.332 & 0.019 & 0.017 & 0.010 \\       
Greek & 75 & 0.331 & 0.027 & 0.024 & 0.010       & 74 & 0.333 & 0.021 & 0.019 & 0.008 \\        
Catalan & 98 & 0.332 & 0.023 & 0.021 & 0.008     & 98 & 0.333 & 0.014 & 0.012 & 0.006 \\      
Portuguese & 88 & 0.331 & 0.024 & 0.023 & 0.009  & 88 & 0.332 & 0.013 & 0.012 & 0.006 \\   
Spanish & 95 & 0.332 & 0.023 & 0.022 & 0.009     & 95 & 0.333 & 0.014 & 0.013 & 0.006 \\      
Persian & 93 & 0.329 & 0.023 & 0.021 & 0.009     & 93 & 0.331 & 0.022 & 0.021 & 0.009 \\     
Czech & 88 & 0.330 & 0.024 & 0.022 & 0.010       & 87 & 0.331 & 0.019 & 0.017 & 0.009 \\        
English & 74 & 0.331 & 0.033 & 0.031 & 0.013     & 75 & 0.333 & 0.023 & 0.021 & 0.010 \\      
Swedish & 74 & 0.329 & 0.028 & 0.026 & 0.011     & 73 & 0.331 & 0.022 & 0.021 & 0.010 \\      
Slovak & 92 & 0.330 & 0.024 & 0.022 & 0.010      & 87 & 0.331 & 0.020 & 0.018 & 0.010 \\       
Russian & 80 & 0.330 & 0.024 & 0.022 & 0.010     & 80 & 0.332 & 0.017 & 0.016 & 0.008 \\      
Italian & 69 & 0.331 & 0.024 & 0.022 & 0.010     & 68 & 0.333 & 0.014 & 0.012 & 0.008 \\      
Bulgarian & 64 & 0.330 & 0.029 & 0.027 & 0.012   & 63 & 0.332 & 0.016 & 0.014 & 0.009 \\    
Hindi & 69 & 0.332 & 0.020 & 0.018 & 0.008       & 69 & 0.331 & 0.012 & 0.010 & 0.007 \\        
Japanese & 44 & 0.330 & 0.021 & 0.020 & 0.010    & 44 & 0.333 & 0.008 & 0.006 & 0.007 \\     
Finnish & 41 & 0.329 & 0.028 & 0.025 & 0.016     & 41 & 0.331 & 0.024 & 0.021 & 0.013 \\      
Basque & 35 & 0.331 & 0.026 & 0.022 & 0.017      & 35 & 0.331 & 0.024 & 0.021 & 0.015 \\       
Romanian & 46 & 0.332 & 0.023 & 0.021 & 0.011    & 46 & 0.333 & 0.012 & 0.010 & 0.008 \\     
Bengali & 18 & 0.322 & 0.034 & 0.028 & 0.026     & 17 & 0.321 & 0.034 & 0.027 & 0.030 \\      
Turkish & 51 & 0.332 & 0.030 & 0.027 & 0.013     & 49 & 0.331 & 0.015 & 0.013 & 0.010 \\     
Estonian & 25 & 0.331 & 0.036 & 0.032 & 0.020    & 25 & 0.332 & 0.032 & 0.028 & 0.019 \\      
Tamil & 40 & 0.333 & 0.023 & 0.020 & 0.011       & 40 & 0.333 & 0.018 & 0.016 & 0.010 \\        
Telugu & 10 & 0.330 & 0.043 & 0.030 & 0.030      & 10 & 0.331 & 0.037 & 0.029 & 0.033 \\       
\end{tabular}
\end{ruledtabular}
\end{center}
\end{scriptsize}
\end{table*}

\section{Discussion}

\label{discussion_section}

We have shown that $E_2[C]$ predicts $C_{true}$ with small error, much better than the baseline. 
The positive results are not surprising given the previous success of $E_2[C]$ predicting crossings on uniformly random trees, where $\Delta_2$ is about $5\%$, i.e. about 6 times smaller than the baseline $\Delta_0$, for sufficiently long sentences \cite{Ferrer2014c}.  
It is also worth noting that $E_2[C]$ behaves well even in the treebanks with the lowest proportion of crossings, where one could argue that grammar would impose the heaviest constraints against crossings. For example, it achieves a particularly low relative error in the Romanian and Japanese Prague treebanks although they contain no or almost no crossings (Table \ref{relative_table}).


From a linguistic standpoint (recall Section \ref{methods_section}), notice that $E_2[C]$ does not achieve its worst performance in languages known for their high word order freedom such as Ancient Greek and Latin (which are also the ones with the highest number of crossings according to Table \ref{relative_table}) based on Stanford dependencies; however, its relative performance worsens for these languages when Prague dependencies are employed. 
In Table \ref{relative_table}, the average $\Delta_2$ with Stanford dependencies indicates that $E_2[C]$ is able to make its best predictions in Estonian and Telugu, two agglutinating languages, with other agglutinating languages like Japanese or Tamil also showing better predictions than average. 
However, this may be an effect of the shorter sentences observed in these languages (Tables 5 and 6 of \cite{Ferrer2017a}) and the tendency of the errors of the predictor to be smaller in sufficiently short sentences (Figure \ref{relative2_figure}).

If we instead look at the table obtained by grouping by sentence lengths (Table \ref{relative_grouping_by_sentence_length_table}), we observe that the predictor is remarkably robust across very dissimilar language types and families. As a representative example, if we focus on Stanford dependencies, the best prediction (average $\Delta_2 = 0.014$) is obtained for Arabic: an Afro-Asiatic, non-agglutinating language whose treebank contains long sentences with a relatively high number of crossings; while the third best (average $\Delta_2 = 0.021$) corresponds to Japanese: a Japonic, agglutinating language with short sentences and little observed crossings. The situation is very similar with Prague annotations, with Japanese exhibiting the best prediction, and Arabic the second best.
These simple and partial linguistic analyses are just reported to illustrate the potential of future linguistic research that explores in more depth the relationship between language traits and annotation criteria on the one hand, and crossings and predictions on the other.

It could be argued that the good predictions of $E_2[C]$ are not surprising at all 
because the syntactic dependency structures that we have analyzed could be the result of some sophisticated apparatus: a complex language faculty or external grammatical knowledge which could have produced, indirectly, a distribution of dependency lengths and vertex degrees that is favorable for $E_2[C]$. 
Then the input with which the predictor yields good predictions, e.g., dependency lengths, would be an indirect result of that complex device.
However, $E_2[C]$ does not require such a device: $E_2[C]$ also makes accurate predictions on uniformly random trees with a small number of crossings \cite{Ferrer2014c}. Therefore, the need of external grammatical knowledge to explain the origins of non-crossing dependencies is seriously challenged. 

The high precision of $E_2[C]$ suggests that the actual number of crossings in sentences might be a side effect of the dependency lengths, which are in turn constrained by a general principle of dependency length minimization (see \cite{Ferrer2013e,Liu2017} for a review of the empirical and theoretical backup of that principle).
A ban on crossings by grammar (e.g., \cite{hudson07,Tanaka97}),
a principle of minimization of crossings \cite{Liu2008a} or a competence-plus \cite{Hurford2012_Chapter3} limiting the number of crossings, may not be necessary to explain the low frequency of crossings in world languages. 


In spite of the arguments in favor of a model predicting crossings based on dependency lengths reviewed and expanded in this article, other factors must be considered. First, chunks, i.e. subsequences of words that work as a unit, could also contribute to explain the scarcity of crossings: the number of crossings has been shown to reduce when chunks are sufficiently small in computer experiments \cite{Lu2016a}. 
Second, it looks difficult to rule out 
some principle of minimization of crossings or planarity constraint. The reason is the positive correlation between crossings and dependency lengths that has been unveiled by this article and previous research combining both theory and experiment (see \cite{Ferrer2014c}, \cite{Ferrer2014f}, \cite{Ferrer2015c} and references therein). The question is: what is the causal force for the scarcity of crossings: (a) a principle of minimization of crossings that explains why dependency lengths are short or (b) a principle of dependency length minimization that explains the scarcity of crossings? \cite{Ferrer2014f}. A temporary solution to this dilemma is straightforward if we are seriously concerned about the construction of a general theory of language that is not only highly predictive but also parsimonious: a theory of language based on (b) is more parsimonious than one based on (a) \cite{Ferrer2014f}. 

From a higher perspective, dependency length minimization follows from 
the now-or-never bottleneck, a fundamental constraint on language processing \cite{Christiansen2015a}, and then the scarcity of crossings could be a further prediction of such a fundamental constraint. The latter would imply that the now-or-never bottleneck and the theory of spatial/geographical networks \cite{Reuven2010a_Chapter8,Gastner2006b,Ferrer2004b,Chung1984} are the key for the development of a parsimonious theory of language.

Despite the focus of this article on language, the article is relevant for research on other spatial networks. As researchers on dependency networks have been assuming that syntactic dependency trees tend to be planar or should be planar \cite{sleator93,Tanaka97,KyotoCorpus,Starosta03,Lee04,hudson07,Ninio2017}, research on infrastructure  networks, e.g. road networks, has been assuming that road networks are planar (see \cite{Viana2013a} and references therein) while indeed crossings in road networks cannot be neglected \cite{Eppstein2008a}. In the domain of infraestructure networks, we could borrow questions that have been formulated for syntactic dependency networks: is the number of crossings actually small? \cite{Ferrer2017a}. Can the number of crossings of real infraestructure networks be explained as a result of pressure to reduce crossings directly or indirectly as a result of some principle of dependency length minimization? (this article). These are questions that may not have the same answer as in syntactic dependency trees and that could be illuminated with extensions or generalizations of the theoretical framework reviewed in this article for the two dimensional continuous case. We hope that our work stimulates further research in the field of spatial networks.


\begin{acknowledgments}

We thank Morten Christiansen for helpful discussions, Wolfgang Maier for comments on an earlier version of this manuscript, and Dan Zeman for help with data conversion.

RFC is funded by the grants 2014SGR 890 (MACDA) from AGAUR (Generalitat de Catalunya) and also
the APCOM project (TIN2014-57226-P) from MINECO (Ministerio de Economia y Competitividad).
CGR has received funding from the European Research Council (ERC) under the European Union's Horizon 2020 research and innovation programme (grant agreement No 714150 - FASTPARSE) and from the TELEPARES-UDC project (FFI2014-51978-C2-2-R) from MINECO.

\end{acknowledgments}




\newcommand{\beeksort}[1]{}

\end{document}